\documentclass[conference]{IEEEtran}
\usepackage{times}

\usepackage[numbers]{natbib}
\usepackage[bookmarks=true]{hyperref}
\usepackage{multicol}
\usepackage{graphicx}
\usepackage{algorithm}
\usepackage[noend]{algpseudocode}
\usepackage{multirow}
\usepackage{amsfonts}
\usepackage{nomencl}
\usepackage{amsmath}
\usepackage{threeparttable}
\usepackage{xcolor}
\usepackage{units}
\usepackage{subcaption}
\usepackage{pbox}
\usepackage{textcomp}

\begin{document}

\clearpage
\begingroup
\onecolumn 
{\huge
\noindent  \textcopyright 2019 IEEE.  Personal use of this material is permitted.  Permission from IEEE must be obtained for all other uses, in any current or future media, including reprinting/republishing this material for advertising or promotional purposes, creating new collective works, for resale or redistribution to servers or lists, or reuse of any copyrighted component of this work in other works.
}

\clearpage
\endgroup
\newpage
\twocolumn

\title{Robust Recovery Controller for a Quadrupedal Robot using Deep Reinforcement Learning
}


\author{
\authorblockN{
Joonho~Lee,
Jemin~Hwangbo,
and~Marco~Hutter}
\authorblockA{
Robotic Systems Lab, ETH Zurich\\
E-mails: \{jolee, jhwangbo, mahutter\}@ethz.ch}
}



\maketitle

\begin{abstract}
The ability to recover from a fall is an essential feature for a legged robot to navigate in challenging environments robustly.
Until today, there has been very little progress on this topic. Current solutions mostly build upon (heuristically) predefined trajectories, resulting in unnatural behaviors and requiring considerable effort in engineering system-specific components.
In this paper, we present an approach based on model-free Deep Reinforcement Learning (RL) to control recovery maneuvers of quadrupedal robots using a hierarchical behavior-based controller.
The controller consists of four neural network policies including three behaviors and one behavior selector to coordinate them.
Each of them is trained individually in simulation and deployed directly on a real system.
We experimentally validate our approach on the quadrupedal robot ANYmal, which is a dog-sized quadrupedal system with 12 degrees of freedom.
With our method, ANYmal manifests dynamic and reactive recovery behaviors to recover from an arbitrary fall configuration within less than \unit[5]{seconds}.
We tested the recovery maneuver more than 100 times, and the success rate was higher than \unit[97]{\%}.
\end{abstract}



\section{Introduction}
\begin{figure}[t]
\centering
      \includegraphics[width=\columnwidth]{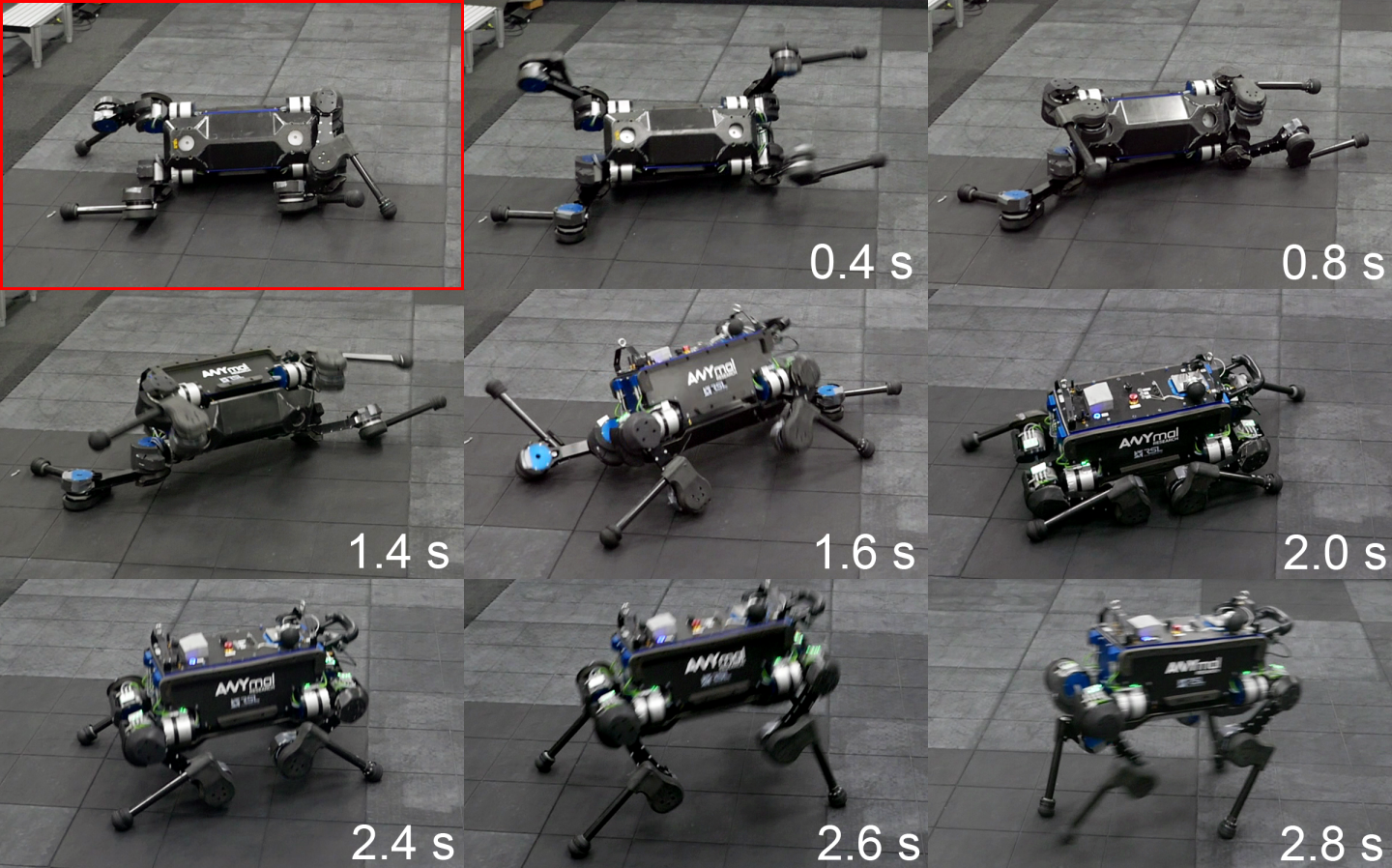}
      \caption{A recovery maneuver of ANYmal. \textbf{(Top left)} ANYmal is initialized at a fall configuration. \textbf{(Top row)} ANYmal swings its legs to gain momentum, \textbf{(Middle row)} pushes the ground to regain the upright and stable posture, and then \textbf{(Bottom row)} stands up and walks.}
      \label{intro}
\end{figure}

In case of a fall, animals show the remarkable ability to recover from any posture by pushing against their surroundings and swinging limbs to gain momentum.
Having similar abilities in legged robots would significantly improve their robustness against failure and extend their applicability in harsh environments.
We address this topic in the present work by developing a control strategy for the robust recovery maneuver of quadrupedal robots.
By recovery maneuver, we mean the maneuver of returning to a typical operating state (standing or walking) from a fall as shown in the Fig.~\ref{intro}.
For such maneuver, the robot needs to produce motions that make good use of the interactions with the ground and swinging motion of the legs while avoiding self-collisions.
Optimization-based methods~\cite{mordatch2012discovery} have a hard time to solve such a task as they usually require analytic dynamic models and often predefined contact sequences~\cite{bellicoso2018dynamic}, which are both hard to find as the system can interact at multiple uncertain contact points or patches. Moreover, all control methods that are based on simplified template models are not valid anymore in such fall configuration.
Existing methods in recovery controllers simplify the problem by using handcrafted control sequences~\cite{semini2015design}, \cite{stuckler2006getting} or using simplified models~\cite{saranli2004model}, or even adding mechanisms such as a tail or extra limbs~\cite{spotMini},~\cite{chen2013hopping}.
Consequently, they exhibit predictable behavioral patterns, which limit their robustness in corner cases (e.g., when a robot's legs get stuck below its base).
They also require a considerable engineering effort.

A promising alternative for generating self-righting behaviors is model-free Deep Reinforcement Learning~(RL).
In this paradigm, an agent interacts with its environment and learns the dynamics and a control policy from the experience.
By using model-free methods, we can generate control policies without any abstraction in modeling complex dynamics.
%
Unfortunately, existing model-free algorithms typically require an excessive number of trial-and-errors to obtain a performant policy, and hence it often becomes impractical to train a policy on sophisticated hardware. 
This is particularly true for dynamic systems like legged robots where bad policies can be fatal.
To overcome these limitations, existing works leverage simulations where one can generate massive data at no cost and with high consistency. 
Very recent and promising research results in the field of legged robotics have demonstrated that learned locomotion policies can be transferred from simulation to reality~\cite{SCIENCE},~\cite{tan2018sim}. 
In order to realize this transfer and overcome the so-called reality gap, it was important to use high-fidelity simulations. 
In \cite{tan2018sim} this was achieved by model parameter estimation, while \cite{SCIENCE} proposed a method to learn parts of the simulated model from real data. 
Beside traditional locomotion, the latter work also demonstrated a sophisticated self-righting policy that can be trained in a few hours in the simulation.
However, it is still challenging to train a single policy that can manifest multiple behaviors including self-righting, standing up, locomotion or others, because the present Deep RL algorithms have limited capability in learning multiple skills. 
It often shows the degradation of the performance in individual tasks in case of learning multiple skills at once (e.g., Multitasker in \cite{berseth2018progressive}).
A simple yet powerful solution to incorporate multiple behaviors is to divide-and-conquer.
In this approach, a control problem is decomposed into several behaviors, and then the behaviors are coordinated either by hand~\cite{Brooks_1986} or learning a rule for behavior selection.
The behaviors and the selection rule can be learned separately~\cite{merel2018hierarchical}, \cite{Liu2017-hd} or simultaneously~\cite{frans2017meta},~\cite{peng2016terrain}.
For separate learning, a high-level behavior selector is trained using a set of pre-learned behaviors.
This is a widely adopted approach in behavior-based robotics~\cite{maes1990learning} and has shown success in high-dimensional continuous control problems recently. \citet{merel2018hierarchical} and \citet{Liu2017-hd} demonstrated human-like behaviors in simulation with a hierarchically structured controller that consists of multiple low-level controllers and an agent to organize them in a task-oriented way.

In this paper we present a hierarchically structured controller consisting only of neural networks that is able to generate complex combined maneuvers like recovering from a fall. 
we build up the complex skill set from individual behaviors including self-righting, standing up, and locomotion.
The self-righting and locomotion behaviors are introduced in the previous work~\cite{SCIENCE}.
We reproduced the self-righting behavior with different action representation and reused existing policy for the locomotion behavior.
These behaviors are trained using Trust Region Policy Optimization (TRPO)~\cite{schulman2015trust} using the simulation framework presented in the existing work~\cite{SCIENCE}. 
We take inspiration from~\cite{merel2018hierarchical} and \cite{Liu2017-hd}, and learn the complex behavior selection subsequently.
Moreover, we introduce a novel learning-based state estimator that is learned in parallel to the behavior selection and which even works in degenerated conditions (i.e., when fallen on the ground and being in a complex, unobservable contact condition).

With these elements, we are able to generate a recovery controller of unprecedented robustness. The system was tested on the quadrupedal robot ANYmal~\cite{hutter2016anymal} in more than 100 trials with a success rate higher than \unit[97]{\%}. Thereby, we showed that our method can cope with all kinds of corner cases for which previous solutions failed.
As the propose controller is not based on any heuristics, it has the potential to be applicable for a wide variety of complex skill sets and hence bring our robots as step closer to their natural counterparts.

\begin{figure}[bt]
\centering
      \includegraphics[width=\columnwidth]{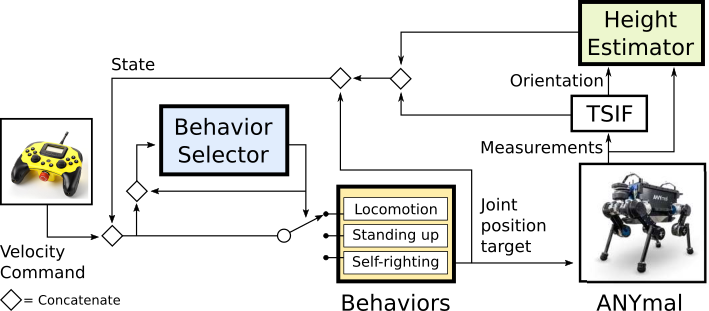}
      \caption{Control architecture for the recovery controller. TSIF refers to the Two State Implicit Filter~\cite{bloesch2018two}.}
      \label{arch}
\end{figure}
\section{Method}
In this section, we first provide an overview of our method and then describe the details of the designs and implementation of each component.
\subsection{Overview}
Fig.~\ref{arch} shows an overview of the control framework.
The behavior selector and three behaviors form a hierarchical behavior-based controller.
Each behavior is a control policy for individual behaviors.
The behavior selector selects the most appropriate behavior for the current situation depending on the recent observations, commands, previously chosen behavior type and the previous action. At each time step, the control policy for the chosen behavior sends commands to the actuators.
The height estimator is a neural network regression module for estimating base height during the deployment on the real system.



We decomposed the control task into three behaviors: self-righting, standing up, and locomotion. This decomposition is to cope with the difficulty of training a single policy that can manifest all of the necessary behaviors.
From our experience, a policy trained to perform multiple tasks shows ungraceful behaviors such as frequent slippages and highly conservative postures.
It also requires a lot more effort to come up with an effective cost function that leads to natural motions because the desired controller has to cover larger state space.
Learning three behaviors separately simplifies the cost function design and enables us to troubleshoot each control policy separately on the real system.

During the deployment on the real system, the estimated states from the Two State Implicit Filter (TSIF)~\cite{bloesch2018two} are used.
Additionally, we used a neural network named height estimator to estimate the base height ($z$-position of the base in the world frame) to resolve the drift issues in linear position.
The definition of the state spaces and the characteristics of TSIF are provided in the section~\ref{statefeatures}.

Each behavior is separately trained and tested on the real robot, whereby training is done only in simulation.
Subsequently, the behavior selector and a height estimator are trained using the set of pretrained behaviors.
The simulated environments for learning consists of the data-driven actuator model~\cite{SCIENCE} and the fast contact solver presented in~\cite{Hwangbo_Lee_Hutter_2018}, which efficiently generates high-fidelity samples.
Trust Region Policy Optimization algorithm~\cite{schulman2015trust} with the Generalized Advantage Estimator~\cite{schulman2015high}~(TRPO+GAE) is used for learning.
Although stochastic policies are used during training, the variances are reduced to 0 during deployment to ensure a more consistent behavior.




\subsection{Feature Selection for the State Spaces}\label{statefeatures}
For each behavior, we select the most representative and reliable set of states as given in table~\ref{stateSpace}.
The existing state estimation framework of ANYmal relies on the Two State Implicit Filter (TSIF)~\cite{bloesch2018two} for the base pose and twist, and angular encoders for the joint states.
TSIF estimates the base twist and the base position in the inertial frame by recursively fusing kinematics and measurements from the Inertial Measurement Unit~(IMU).
It makes use of the positions of feet touching the ground to incorporate kinematic contact constraints.
When a foot slips or all four feet lose contacts with the ground, which are likely to happen when ANYmal falls, the estimated base states from TSIF becomes unreliable as the position and linear velocity drift over time (see section~\ref{sim2Real}).
To maximize reliability, the linear position and linear velocity are excluded from the state space of the self-righting controller.

A unit vector pointing in the direction of the gravity expressed in the base frame~($e_g$) is used to represent the orientation of the base because an IMU can observe only this two dimensional subspace of the orientation (yaw angle is unobservable). 
The $x$, $y$ positions of the base in the inertial frame are excluded from the state spaces because they are always unobservable with IMU~\cite{bloesch2013state}.

On the other hand, the $z$ position of the base (base height) can be accurately estimated using kinematics while walking.
We kept this state in the locomotion policy as it is critical for fast learning and estimate the height using a new estimation network presented in section~\ref{sec:heightestimation}.




\begin{table}[bt]
\centering
\vspace*{0.3cm}
 \begin{tabular}{|l|l|}
 \hline
 Function & Data\\ 
  \hline
  \multirow{6}{*}{Self-Righting Policy} 
                 &Gravity vector ($e_g$) \\
                 &Base angular velocity in body frame ($\omega^B_{IB}$)\\
                 &Joint positions ($\phi_j$)\\
                 &Joint velocities ($\dot{\phi_j}$)  \\
                 &History of joint position error \& velocity \\
                 &Previous joint position targets ($a_{t-1}$) \\
 \hline
 \multirow{2}{*}{Standing Up Policy} 
                &Base linear velocity in body frame ($v^B_{IB}$)\\
                 &State space of the Self-Righting policy\\
 \hline  
 \multirow{3}{*}{Locomotion Policy}
                 &Velocity commands \\
                 &Estimated base height ($h_{e}$)\\
                 &State space of the Standing up policy\\
 \hline
 \multirow{2}{*}{Behavior Selector} 
                 &Previous action (one-hot vector) \\
                 &State space of the Locomotion policy\\
 \hline 
 \multirow{4}{*}{Height Estimator} 
                 &Gravity vector ($e_g$) \\
                 &Joint positions \\
                 &Joint velocities  \\
                 &History of joint position errors \& velocities\\
 \hline 
 \multirow{2}{*}{Actuator Model} 
                 &Desired joint position\\
                 &History of joint position errors \& velocities\\
 \hline
 \end{tabular}
  \caption{Definition of the State Spaces}
 \label{stateSpace}
\end{table}

\subsection{Cost terms}
\label{costTerms}
The cost terms are presented in the table~\ref{costTable}.
We defined cost functions for each behavior by a linear combination of cost terms.
For joint position cost, we used the minimum angle difference between two angular positions denoted by $d_{\phi}(\cdot, \cdot): \mathbb{R} \times \mathbb{R} \rightarrow [0, \pi]$.

It is important to define a bounded cost function because otherwise an agent often finds it more rewarding to terminate than to explore its environment.
To this end, we used a logistic kernel function $K: \mathbb{R} \rightarrow [-0.25, 0)$ defined as
$
K(e|\alpha) = - 1/{(e^{\alpha e} + 2 + e^{-\alpha e})}, \quad \alpha \in \mathbb{R}_{>0}
$
where $e$ represents an error term.
This kernel comes handy in tuning because it enables us to leverage relative importance between different cost terms and it enables us to adjust an agent's sensitivity to $e$ by adjusting $\alpha$.

\begin{table}[bt]
\centering
\vspace*{0.3cm}
 \begin{tabular}{|ll|}
 \hline
\multicolumn{2}{|c|}{Symbols}\\
\hline
$\dot{\phi}_{jslim}$ & maximum joint speed \\
$I_{c}$ & index set of the contact points \\
$I_{c,f}$ & index set of the foot contact points\\
$I_{c,in}$ & index set of the self-collision points\\
$i_{c,n}$ & impulse of the $n$th contact\\
$g_i$ & gap function of the $i$th contact\\
$p_{f,i}$ & position of the $i$th foot\\
$\tau$ & vector of joint torques\\
$\lvert \cdot \rvert$ & cardinality of a set or $l_1$ norm\\
$\lvert\lvert \cdot \rvert\rvert$ & $l_2$ norm\\
$\hat{\cdot}$ & target value \\
\hline
\multicolumn{2}{|c|}{Cost Terms} \\
  \hline
  Angular velocity & $c_{\omega} =  K(\lvert \omega^B_{IB} - \hat{\omega}^B_{IB}\rvert, \alpha_a)$\\[0.5ex]
 Linear velocity & $c_v =  K(\lvert v^B_{IB} - \hat{v}^B_{IB}\rvert, \alpha_l)$\\[0.5ex]
  Height  & $c_h =  1.0  $ if $h < 0.35$ , otherwise 0  \\[0.5ex]
 Joint position & $c_{jp} = d_{\phi}(\phi_{j}, \hat{\phi_{j}})$\\[0.5ex]
 Orientation &
    $c_o = \lvert \lvert [0,0,-1]^T - e_{g} \rvert \rvert$ \\[0.5ex]
  Torque & $c_{\tau} =  \lvert\lvert \tau \rvert\rvert^2$ \\[0.7ex]
Power & $c_{pw} = \sum_{i=0}^{12} max(\dot{\phi}_{j,i} \tau_{i}, 0)$ \\[0.7ex]
Joint acceleration & $ c_a = \sum_{i=0}^{12}  \lvert\lvert \ddot{\phi_i} \rvert\rvert ^2$ \\[0.7ex]
 Joint speed & $c_{js} =  \sum_{i=0}^{12} max(\dot{\phi}_{jslim}- \lvert\phi_{i}\rvert, 0)^2$ \\[0.7ex]
Body impulse & $c_{bi} = \sum_{n \in I_{c} \backslash I_{c,f} }||i_{c,n}|| / (\lvert I_{c} \rvert - \lvert I_{c,f} \rvert)$ \\[0.7ex]
Body slippage & $c_{bs} = \sum_{n \in {I_{c}}}||v_{c,n}||^2 / \lvert I_{c} \rvert$ \\[0.7ex]
Foot slippage & \pbox{\columnwidth}{$c_{fs} =  \sum \lvert\lvert v_{f,i} \rvert\rvert $ \\ \hspace*{2.0cm} $\forall i  \; s.t. \; g_i = 0, i \in I_{f,c}$ }\\[0.7ex]
Foot clearance & \pbox{\columnwidth}{$c_{fc} = \sum (p_{f,i} - 0.07) ^ 2 \lvert\lvert v_{f,i} \rvert\rvert$ \\ \hspace*{2.0cm} $ \forall i  \; s.t. \; g_i > 0, i \in I_{f,c}$ }\\[0.7ex]
Self collision & $c_{cin} = \lvert I_{c,in} \rvert$\\[0.5ex]
Action difference & $c_{ad} =  \lvert\lvert a_{t - 1} - a_{t}\rvert\rvert^2$\\
\hline
 \end{tabular}
  \caption{Cost Terms}
 \label{costTable}
\end{table}

\subsection{Behaviors}
The policies for self-righting, standing up and locomotion are individually trained to achieve different tasks.

\subsubsection{Tasks}
 Each behavior is learned based on a different cost function, initial state distributions, and termination conditions.
 
\begin{figure}[t]
\centering
    \begin{subfigure}{0.4\columnwidth}
   \centering
      \includegraphics[width=\columnwidth]{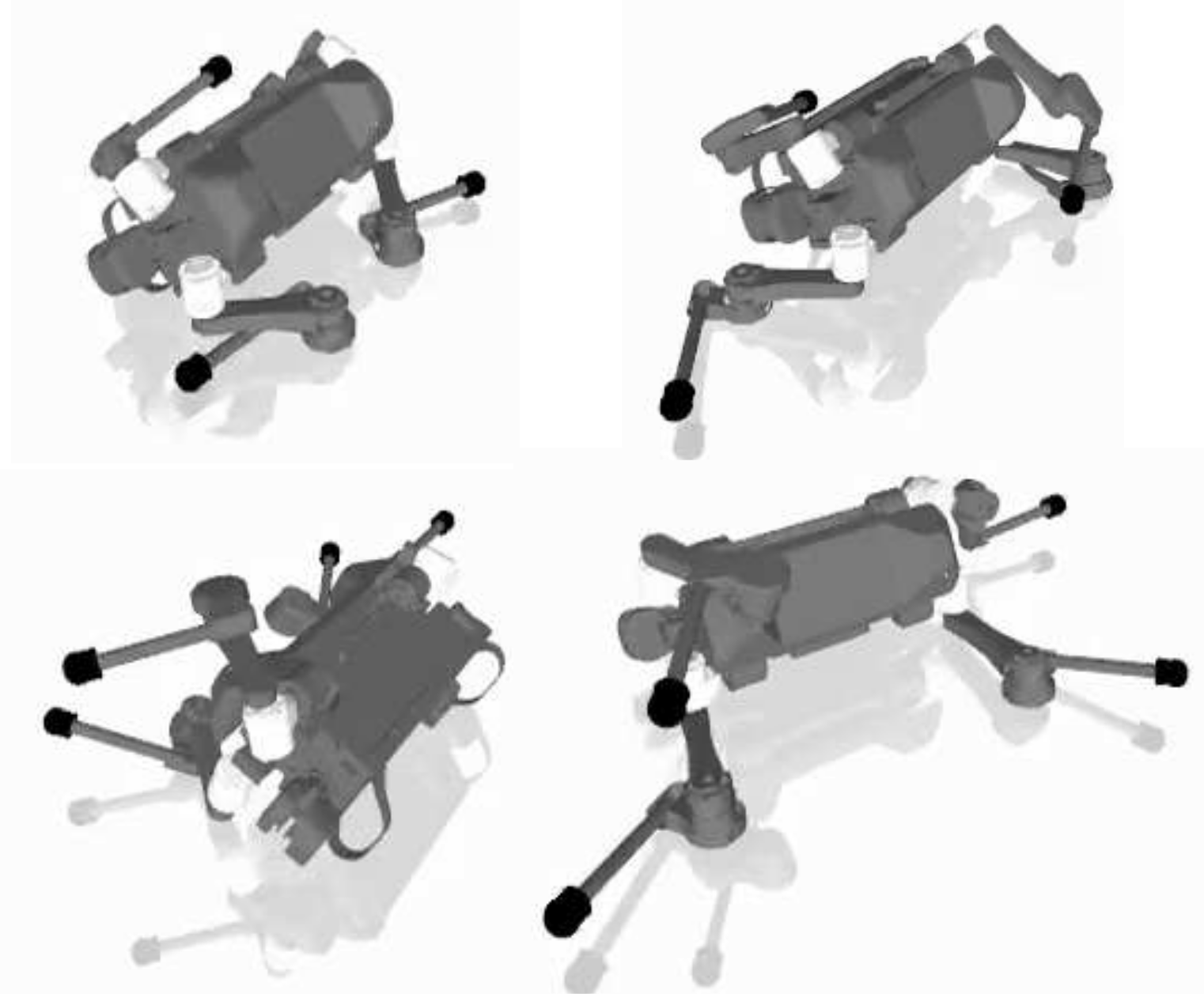}
    \caption{}
      \label{initialStates}
    \end{subfigure}
    ~
  \begin{subfigure}{0.4\columnwidth}
    \centering
      \includegraphics[width=\columnwidth]{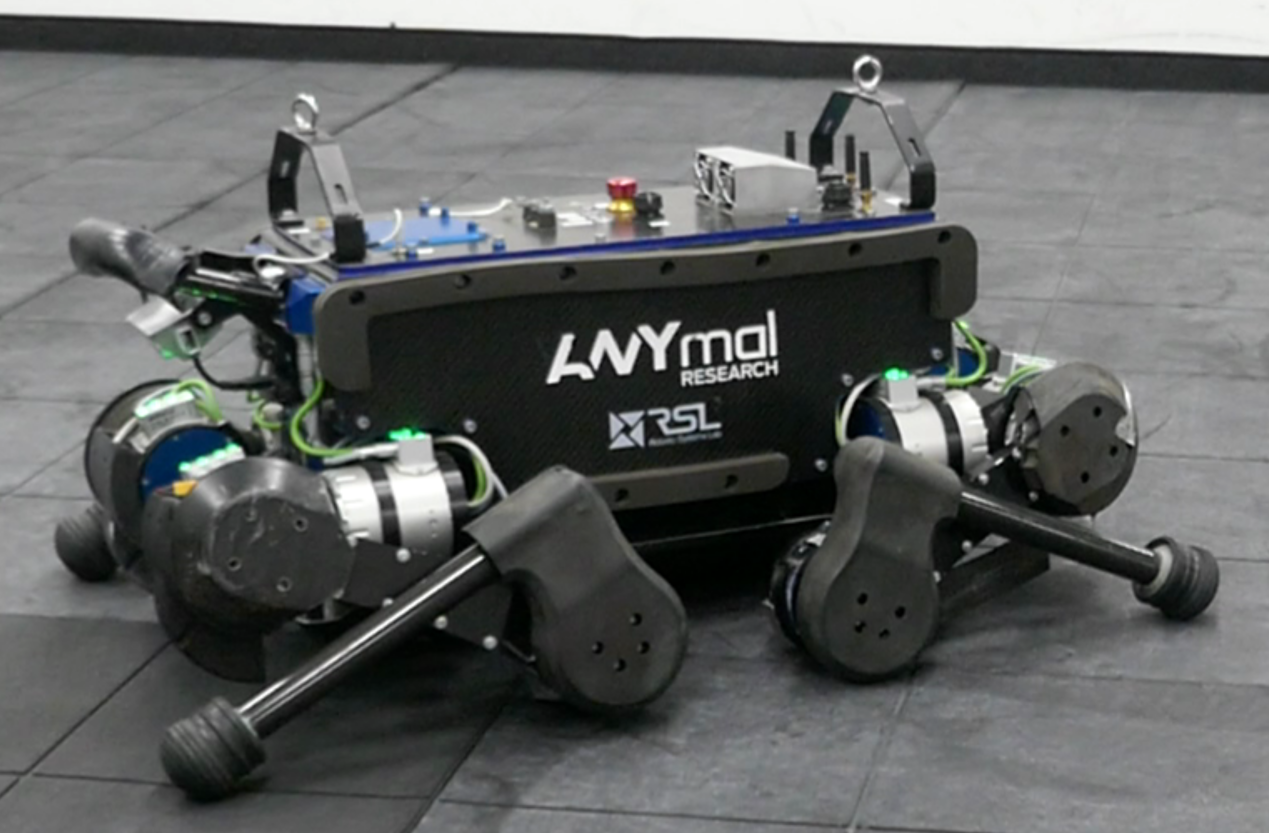}
       \caption{}
      \label{flipTarget}
    \end{subfigure}
          \caption{(\textbf{a}) Sampled initial states and (\textbf{b}) the target configuration of the self-righting task.}
\end{figure}

\begin{itemize}
    \item \textbf{Self-righting behavior} is to regain upright base pose from an arbitrary configuration (Fig.~\ref{initialStates}) and re-position joints to the sitting configuration (Fig.~\ref{flipTarget}) which is designed such that ANYmal has all feet on the ground for a safe stand-up maneuver. The cost function is defined as 
$  k_{o} c_{o} + k_{jp} c_{jp} + k_{a} c_a + k_{bi} c_{bi} + k_{bs} c_{bs} + k_{c,in} c_{c,in} + k_{ad} c_{ad} + k_{jslim} c_{jslim} + k_{\tau} c_{\tau} $ , where $k_{(\cdot)}$ is a scaling factor.
Each cost terms are explained in table.~\ref{costTable}.
    The weight for the orientation cost ($k_{o}$) was set to be the highest such that ANYmal recovers up-right base pose as soon as possible. The magnitudes of contact impulses are penalized to avoid violent motions.
    The joint accelerations are penalized to generate smooth motions.
    
    \quad To sample initial states for training in simulation, we dropped ANYmal from 0.5 m above the ground with random joint positions.
    The termination condition is only the time limit.
    
    \item \textbf{Standing up behavior} is to stand up from arbitrary sitting configurations.
    The cost function is similar to the cost of the self-righting but a height cost is additionally introduced: $ k_{jp} c_{jp} + k_{o} c_{o} + k_{h} c_h + k_{a} c_a + k_{ad} c_{ad} + k_{jslim} c_{jslim} + k_{\tau} c_{\tau}$.
    The target joint configuration is the standing configuration.
    
    \quad We use the same strategy for sampling initial states as the self-righting task but with a near-upright pose and do not specify any termination condition except for the time limit.
    
    \item \textbf{Locomotion behavior} is to follow a given velocity command composed of desired forward velocity, lateral velocity, and turning rate (or yaw rate) which are sampled from the uniform distribution, $\unitfrac[U(-1,1)]{m}{s}$, $\unitfrac[U(-0.4,0.4)]{m}{s}$, and $\unitfrac[U(-1.2,1.2)]{rad}{s}$ respectively. They are defined concerning the capabilities of an existing controller~\cite{bellicoso2018dynamic}. The cost function penalizes the velocity tracking errors ($ c_{\omega}$ and $c_v$), foot motions ($c_{fc}$ and $c_{fs}$), and constraint violation. 
    It is defined as $ k_{\omega} c_{\omega} + k_{v} c_v + k_{o} c_o + k_{fc} c_{fc} + k_{fs} c_{fs} + k_{ad} c_{ad} + k_{jslim} c_{jslim} + k_{\tau} c_{\tau}$.
    
    \quad The initial states are sampled from a multivariate Gaussian distribution centered at the standing configuration.
    An episode terminates when the joint limit is violated or ANYmal falls.
\end{itemize}


\subsubsection{Action Representation}
The output of a policy for a behavior is a 12-dimensional real-valued vector and each dimension is mapped to a position target of the low-impedance Proportional-Derivative (PD) controllers running at each joint actuator.
\citet{peng2017learning} showed that using this type of action representations for learning motor skills achieves better final performance, faster learning, and higher robustness compared to other representations such as target joint torques and target joint velocities.

The output of a policy network is mapped to joint position targets differently depending on the task.
For locomotion, the desired joint position~$\phi_d$ is defined as $\phi_d = k o_t + \phi_n$ where $k$ is a scaling parameter, $o_t$ is the output, and $\phi_n$ is a nominal joint configuration (standing).
It is designed such that the distribution of the target positions has a standard deviation of approximately 1 and mean at the nominal configuration at the beginning of training.
It accelerates the learning because the agent explores trajectories near the standing configuration more frequently.
For the self-righting and the standing up, we define $\phi_d = k o_t + \phi_t$, where $\phi_t$ is a vector of current joint positions. 
It promotes exploration in joint spaces and results in faster learning of self-righting.

\subsubsection{Architecture}
The policies are parameterized by a two-layered feed-forward neural network with tanh units in the hidden layers.
The self-righting and standing-up policies have 128 units in each hidden layer and the locomotion policy has 128 and 256 units, respectively.

\subsubsection{Training}
Each policy is trained separately with TRPO+GAE.
For the natural and smooth motions, we penalized joint torque, velocity, acceleration, and action difference.
We employed Curriculum Learning (CL) in a way that these terms are near zero at the first iteration and increase as the training proceeds~\cite{SCIENCE}; otherwise a learning agent converged to a local minima where it stops moving.

\subsection{Behavior Selector}
The behavior selector learns to determine which behavior to execute depending on the current situation.

\subsubsection{Task}
A behavior selector has to choose an appropriate behavior such that ANYmal returns to a nominal operating state every time it loses balance.
By a nominal operating state, we mean states where it can locomote.
To this end, the cost function is defined as $k_{\omega}c_{\omega} + k_v c_v + k_o c_o + k_h c_h + k_{pw} c_{pw} + k_{ad} c_{ad} + k_{jslim} c_{jslim} + k_{\tau} c_{\tau}$, which is similar to that of the locomotion task.
Additionally,
the use of $c_{pw}$ makes the resulting behaviors power efficient, while
$c_{ad}$ and $c_{\tau}$ ensure smooth transitions between different behaviors.

The initial states are sampled from the initial state distributions of a randomly selected behaviors and the termination condition is the same as that of self-righting.

\subsubsection{State and Action Spaces}
The state space of the behavior selector consists of the union of the state spaces of behaviors and the index of previously chosen behavior.

A behavior selector maps a state $s$ to a categorical distribution over the behaviors.
It is denoted as $\pi_{\theta}(a|s)$ with $a \in \{ 0, 1, 2\}$ and the output is a three dimensional real-valued vector $\{ p_0, p_1, p_2 \}$. Each dimension represents the probability for choosing each behavior.

\begin{algorithm}
\caption{Training Behavior Selector}
\begin{algorithmic}
\State Initialize $\theta$, $\psi$ randomly
\For{$i = 0, 1, ..., N$}
\For{$t = 0, 1, ..., T$}
\If{$i > N_w$} \Comment{$N_w$ = Warm-up period}
\State Use the estimated height $h_{\psi}(s_t)$
\EndIf
\State Sample action $a_t \sim \pi_{\theta}(a|s_t)$
\State Excute the corresponding behavior
\State Collect state $s_t$, action $a_t$, and reward $r_t$
\State Collect true height $h_t$.
\State Append a $s_t$-$h_t$ pair into the replay memory
\EndFor
\State Sample $K$ pairs from the replay memory
\State Update $\psi$ by minimizing $\sum_{j=0}^{K} \lvert\lvert h_j - h_{\psi}(s_j) \rvert\rvert^2$
\State Update $\theta$ using TRPO~\cite{schulman2015trust}
\EndFor
\end{algorithmic}
\label{algo}
\end{algorithm}
\subsubsection{Architecture}
The behavior selector is parameterized by a two-layered feed-forward neural network with 128 tanh units in the hidden layers. 
Softmax function is used at the output such that $\sum_{i = 0}^{3} \pi_{\theta} (i|s) = 1$ for any $s$.
The height estimator is of the same structure but without Softmax. Moreover it uses the softsign unit  which is computationally more efficient than tanh. 

\subsubsection{Training}
We use a set of pre-trained behaviors, which are regarded as a part of the environment, to train the behavior selector using TRPO+GAE. The height estimator, which is explained in section~\ref{sec:heightestimation}, is trained concurrently as outlined in the algorithm~\ref{algo}.
The reasoning behind this strategy is to match their state visitation frequency.


\subsection{Height Estimation}\label{sec:heightestimation}
The estimated base height becomes unreliable when ANYmal falls.
We could observe huge errors from the base position estimates during the experiments when ANYmal lies on the ground, which can lead to undesired behaviors.
To resolve this issue, we trained a neural network to estimate the base height.
It is denoted as $h_{\psi}$ with a set of parameters $\psi$.
The output is calculated from the body orientation (IMU) and joint positions (encoders), which are states that do not have a drift issue.
The base height can be computed using forward kinematics under an assumption that ANYmal is on the flat ground and the geometric properties of the links are known.

The linear velocity estimate also had the same issue but the errors were not significant.

\subsection{Handcrafted Behavior Selector}\label{FSMMethod}
\begin{figure}[h]
\centering
      \includegraphics[width=\columnwidth]{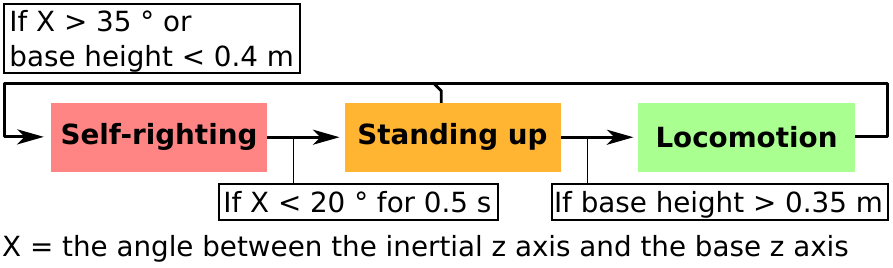}
      \caption{FSM for behavior selection. 
      }
      \label{FSM}
\end{figure}

We introduce a traditional approach that we considered before using the neural network behavior selector. 
Finite State Machine (FSM) is a widely adopted method for controlling hybrid systems.
It is defined by a set of states and transitions between them, which are usually designed by a domain expert.
The proposed controller can be seen as an FSM if we regard behaviors as states and the behavior selector as a learned transition rule.
As the task is straightforward, we could handcraft it by going through trial-and-error (Fig.~\ref{FSM}).
To maximize the success rate, it waits for TSIF to converge for \unit[0.5]{seconds} after it conducts the self-righting behavior. 

\begin{figure}
\centering
      \includegraphics[width=\columnwidth]{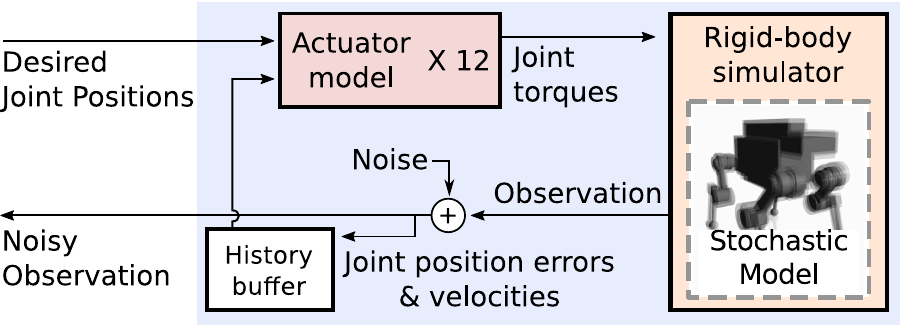}
      \caption{Simulation for ANYmal.}
      \label{sim}
\end{figure}

\begin{figure*}
\centering
    \includegraphics[width=\textwidth]{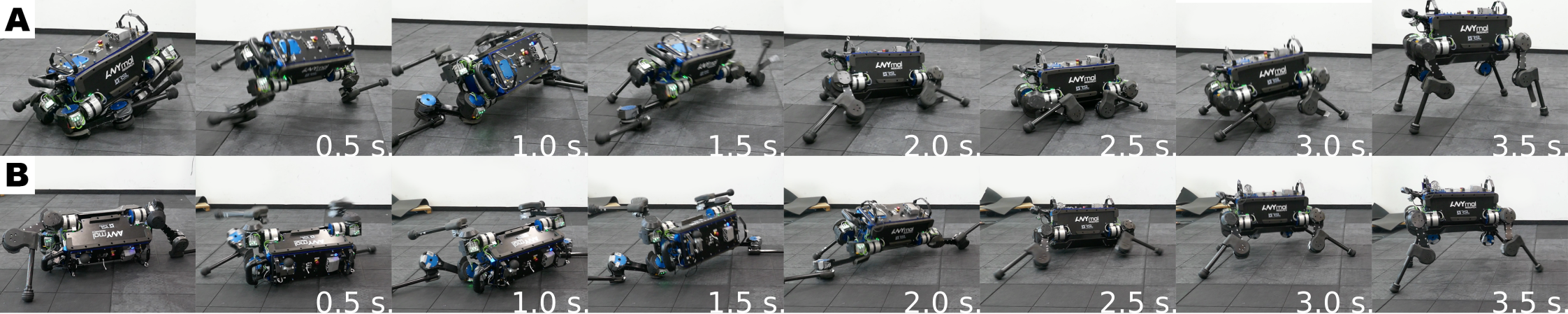}
    \caption{ANYmal recovering from arbitrary configurations.}
    \label{rightings}
\end{figure*}

\begin{figure*}
\centering
    \includegraphics[width=\textwidth]{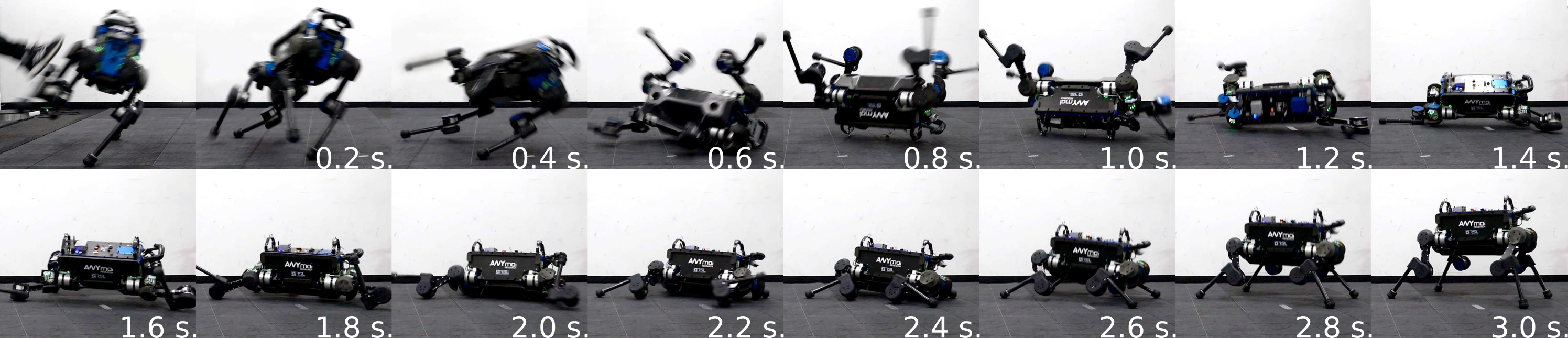}
    \caption{ANYmal reacting to a kick.}
    \label{kick_stand}
\end{figure*}

\subsection{Simulating ANYmal}
The structure of our simulator is depicted in Fig.~\ref{sim}.
It has actuator models and stochastic components that are designed to account for modeling errors and to robustify the resulting policies~\cite{SCIENCE}.
We used RaiSim~\cite{Hwangbo_Lee_Hutter_2018} as rigid-body simulator together with learned networks that represent the actuator dynamics.

\subsubsection{Randomized Physical Properties}
To make the solution robust against modeling errors and to avoid tedious parameter estimation for each link as done in~\cite{tan2018sim},
we directly use physical properties computed from the CAD model including link lengths and inertial properties, but randomize the simulation to overcome modeling errors. 
It has been shown in several works that randomization enhances the robustness and increases the success rate of a sim-to-real transfer~\cite{SCIENCE}, \cite{tan2018sim}, \cite{hwangbo2017control}.
The link masses are randomized with additive noises up to 10 \% of the original value, and the COM of the base is randomly translated up to \unit[3]{cm} in $x$, $y$, $z$ directions respectively every episode.

We approximated the collision geometry of ANYmal with collision primitives such as a box, a cylinder, and a sphere.
The positions and shapes of the these collision bodies are also randomized.
The coefficient of friction between the objects is uniformly sampled between 0.8 to 2.0 every time step because we cannot accurately simulate the material properties.

\subsubsection{Actuator Model}
ANYmal's joints are Series Elastic Actuators~(SEA)~\cite{pratt1995series}.
An SEA consists of multiple components including a spring, gears, encoders, and an electric motor, which results in highly complex dynamics.
It is essential to simulate actuators accurately and fast to efficiently train a sim-to-real transferable policy because it substantially improves the simulation accuracy.
Developing an analytic model requires a large number of parameters to be estimated or assumed to be accurately provided by a manufacturer and thus often results in an inaccurate model~\cite{gehring2016practice}. Instead, we use a data-driven model introduced in~\cite{SCIENCE}.
The actuator model is a neural network that outputs a joint torque conditioned on the position command and a history of joint position errors and velocities.
It is a two-layered feed-forward neural network with 32 softsign units and trained with real data.

\subsubsection{Additive Noise to the Observation}
To replicate the noisy observation from the real robot, we added up to \unitfrac[0.2]{m}{s} noise to the linear velocity, \unitfrac[0.25]{rad}{s} to the angular velocity, \unitfrac[0.5]{rad}{s} to the joint velocities, and \unit[0.05]{rad} to the joint positions during training in simulation.
Additionally, in order to replicate the behavior of TSIF, we increase the magnitude of the noise for the linear velocity and position of the base when all four legs lose contact.

\subsection{Implementation Details}

We used the Robotics Artificial Intelligence~\cite{hwangbo2017control} framework together with the rigid-body simulator~\cite{Hwangbo_Lee_Hutter_2018} , which are both written in C++.

Temporal attributes of an RL task such as the control frequency, the time limit, and the discount factor are regarded as hyper-parameters.
The policies for self-righting, standing up, and locomotion run at \unit[20]{Hz}, \unit[100]{Hz}, and \unit[200]{Hz} respectively and the behavior selector runs at \unit[50]{Hz}.
The height estimator runs synchronously with TSIF, which runs at \unit[400]{Hz}.
When ANYmal switches to a different behavior, the output of the chosen behavior is computed immediately.
As a result, the time step of the self-righting is often not preserved because it runs at the lowest frequency.

As the policies require a history of joint measurements as input, we implemented a history buffer that saves states for \unit[0.05]{seconds} in 400 Hz.
ANYmal waits in freeze mode to fill the history buffer before running the controller.

\section{Result and Discussion}
The experimental results are provided in this section.
The behaviors for standing-up and locomotion are not assessed in detail in this paper. We refer the readers to~\cite{SCIENCE} for a comprehensive analysis of the locomotion behavior.

\begin{figure*}
\centering
    \includegraphics[width=\textwidth]{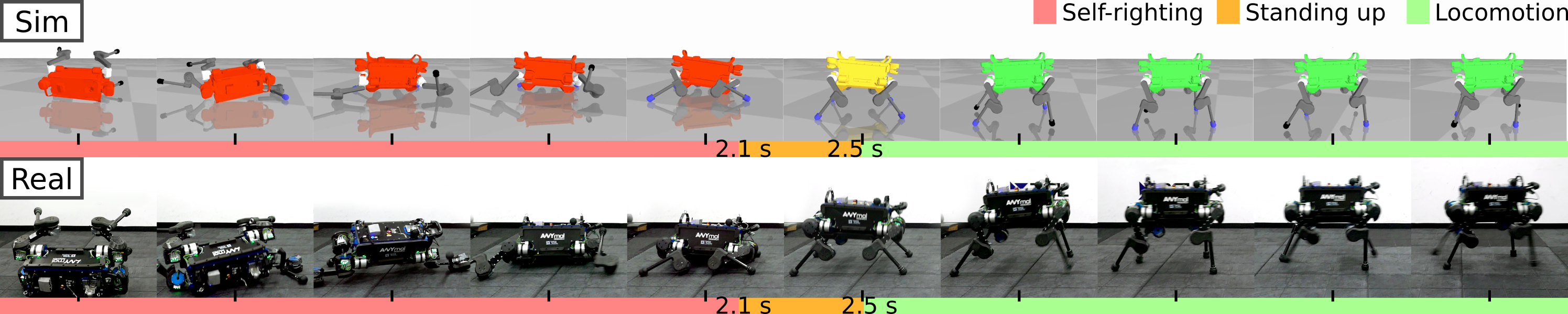}
    \caption{Comparison between simulation and experiments. 
    ANYmal trots in \unit[1]{m/s} after standing up. The snapshots are taken every 0.5 seconds and the color bars represents the active behavior.
    The numbers represent the time of the switching and rounded to the first decimal place.
    (\textbf{Sim}) Simulated ANYmal with the initial state obtained from the experiment.
    (\textbf{Real}) Deploying the recovery controller on the real robot.
    We used the same noises and randomized dynamic properties as in simulation and did not hand-tune any number to matched the motions of the simulation and the experiment.
    }
    \label{snapshots}
\end{figure*}

\subsection{Robustness}
We conducted two kinds of experiments to verify the robustness of the policies.
Firstly, we started the proposed controller while ANYmal lies on the ground at an arbitrary configuration.
We tested 50 different configurations and the self-righting policy could recover up-right base pose within 5 seconds in all experiments.
ANYmal can recover even when its legs are stuck beneath its base (Fig.~\ref{rightings}.A).
It flips to its side to free the legs and then quickly regains up-right base pose.
It recovers when its base is almost upside-down (Fig.~\ref{rightings}.B).

Secondly, we applied external disturbances while ANYmal is walking or standing.
An example is provided in Fig.~\ref{kick_stand}.
It smoothly switches to self-righting behavior or standing up behavior without any noticeable delay.

Both experiments are conducted 50 times and ANYmal fell more than 100 times in total. The recovery maneuver failed only three times. Consequently, The success rate was higher than \unit[97]{\%}.
Self-righting failed when a joint position $\geq 2\pi$.
It fails because a self-righting policy hardly experiences such a high position during the training.
This is very unlikely to happen when ANYmal falls while walking and can be easily fixed by applying modulo operation to the joint positions with 2$\pi$. 


\subsection{Comparison to Simulation}\label{sim2Real}
To qualitatively analyze the accuracy of the simulation, we ran the same controller with the same initial state and velocity command.
As shown in Fig.~\ref{snapshots}, the switch in behaviors appears in simulation (top) and reality (bottom) almost at the same time, and the behaviors are visually identical (Fig.~\ref{snapshots}).


\begin{figure}[bt]
\centering
 \includegraphics[width=1.0\columnwidth]{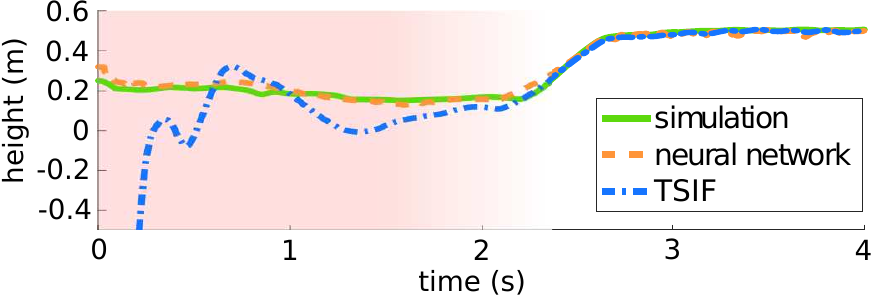}
 \caption{Estimated base height from different sources.
Self-righting policy runs in the shaded region.
 (\textbf{neural network \& TSIF}) Output of the height estimator network and TSIF during the experiment in Fig.~\ref{snapshots}.Real.
  (\textbf{simulation}) Computed height from the simulation in Fig.~\ref{snapshots}.Sim.
  }
 \label{heigtCompare}
\end{figure}
\begin{figure}[bt]
\centering
        \includegraphics[width=\columnwidth]{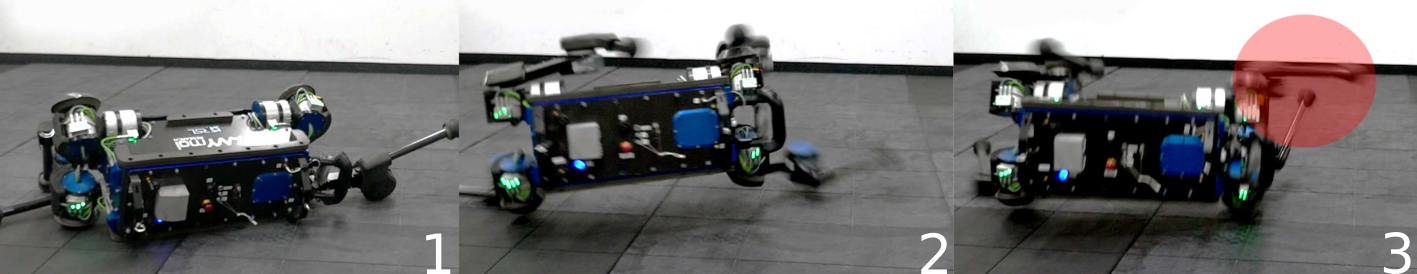}
    \caption{ANYmal showing undesirable behavior without the height estimator. 
    ANYmal jumps up and hits itself (red circle). The behavior selector switches to locomotion policy while ANYmal is lying on the ground.}
    \label{selfHit}
\end{figure}

The estimated heights for the manuever above is provided in Fig.~\ref{heigtCompare}.
Unfortunately, due to a lack of motion capture system, we could not measure the height in the experiment but have only data from simulation.
The output of the TSIF shows a huge error when ANYmal lies on the ground (as shown by the initial value) and fluctuates during the self-righting maneuver.
On the other hand, the output of the neural network is stable and accurately matches the simulated data (the RMS error is less than \unit[1]{cm}).
When deployed without the height estimator, the error in the base height sometimes results in an undesirable behavior switch as shown in the Fig.~\ref{selfHit}.

\subsection{FSM behavior selector}

\begin{figure}
    \centering
   \includegraphics[width=\columnwidth]{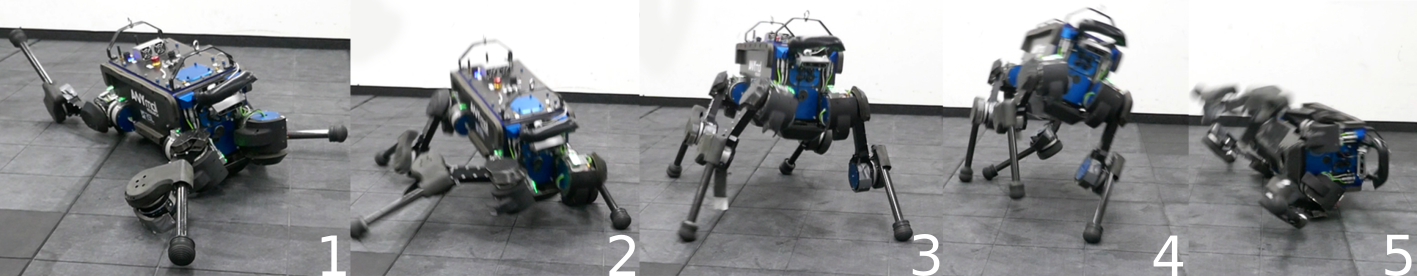}
    \caption{A corner case of our FSM. It switches to standing up policy at a bad timing and falls.}
    \label{fsmFail}
\end{figure}

We discuss the FSM behavior selector introduced in section~\ref{FSMMethod}.
The simplicity of FSM did not allow us to capture corner cases (even with significant tuning in simulation and on the real system) as the one shown in Fig.~\ref{fsmFail}.
Besides, the transitions between the behaviors are often unsmooth.  
The FSM behavior selector is still robust enough to be used in the field.
However, we did not examine the success rate of it because the result can be manipulated if we experiment with the corner cases more.
The performance can be improved by adding more states and transitions. For example, the corner case in Fig.~\ref{fsmFail} can be resolved if we check joint positions before standing up.
However, the fundamental problem is that it requires a number of design iterations and experiments on the real robot.


\section{Conclusion} 
\label{sec:conclusion}
This paper presents a hierarchically structured controller for the quadruped robot ANYmal that can autonomously recover from a fall and locomote on flat terrain.
The control task is decomposed into three behaviors: standing, self-righting and walking.
This strategy made it easier to design well-defined RL tasks and troubleshoot on the real robot.
The policies for behaviors are individually trained to achieve a distinct behavior, and a behavior selector is trained to coordinate them. Additionally, a height estimator is learned, which turned out to be a crucial part for reliable maneuvers.
All the policies are trained in simulation and deployed on the real robot.

The proposed controller exhibits dynamic recovery maneuvers involving multiple ground contacts and the resulting motions are consistent with the simulated ones.
ANYmal can recover from multiple random fall configurations within 5 seconds and switches seamlessly between three behaviors in response to disturbances.
The robustness of our recovery controller is evaluated by testing the controller more than 100 times on the physical system.


The current limitation of the proposed method is that it has only been trained and tested on flat ground and it will fail in case of large inclinations or rough ground. We plan to overcome these limitations in future work by randomizing the environment in simulation and by estimating its properties.

\newpage
\bibliographystyle{plainnat}
\bibliography{/references}  

\addtolength{\textheight}{-12cm}   

\end{document}